\documentclass[runningheads]{llncs}
\usepackage{graphicx}
\usepackage{times}
\usepackage{epsfig}
\usepackage{amsmath}
\usepackage{amssymb}
\usepackage{booktabs} 
\usepackage{enumerate}
\usepackage{adjustbox}
\usepackage{indentfirst}
\usepackage{array}
\usepackage{multirow}
\usepackage{float}
\usepackage{hhline}
\usepackage{bm}
\usepackage{subfigure}

\usepackage{fancyhdr}

\begin{document}

\title{A Performance Evaluation of Loss Functions for Deep Face Recognition}

\author{Yash Srivastava \and Vaishnav Murali \and Shiv Ram Dubey}

\institute{Computer Vision Group,\\ Indian Institute of Information Technology, Sri City, Chittor, A.P., India\\
\email{\{srivastava.y15, murali.v15, srdubey\}@iiits.in}}

\maketitle
\thispagestyle{fancy}
\fancyhf{}
\lhead{Accepted in NCVPRIPG 2019 Conference}

\begin{abstract}
Face recognition is one of the most widely publicized feature in the devices today and hence represents an important problem that should be studied with the utmost priority. As per the recent trends, the Convolutional Neural Network (CNN) based approaches are highly successful in many tasks of Computer Vision including face recognition. The loss function is used on the top of CNN to judge the goodness of any network. In this paper, we present a performance comparison of different loss functions such as Cross-Entropy, Angular Softmax, Additive-Margin Softmax, ArcFace and Marginal Loss for face recognition. The experiments are conducted with two CNN architectures namely, ResNet and MobileNet. Two widely used face datasets namely, CASIA-Webface and MS-Celeb-1M are used for the training and benchmark Labeled Faces in the Wild (LFW) face dataset is used for the testing.
\end{abstract}

\section{Introduction}
Unconstrained face recognition is one of the most challenging problems of computer vision. With numerous use cases like criminal identification, attendance systems, face-unlock systems, etc., face recognition has become a part of our day to day lives. The simplicity of using these recognition tools is one of the major reasons for its widespread adoption in industrial and administrative use. Many scientists and researchers have been working on various techniques to obtain an accurate and robust face recognition mechanism as its use will increase exponentially in coming years.

In 2012, the revolutionary work presented by Krizhevsky et al. \cite{alexnet} with Convolutional Neural Networks (CNN) was one of the major breakthroughs in the image recognition area and won the ImageNet Large Scale Challenge. 
Various CNN based approaches have been proposed for the face recognition task in the past few years. Most of the techniques dealt with including all the complexities and non-linearities needed for the problem and thus obtaining more generalized features and achieving state-of-the-art accuracies over major face datasets like LFW \cite{lfw}, etc. 
Since 2012, many deep learning based face recognition frameworks like DeepFace \cite{deepface}, DeepID \cite{deepid2}, FaceNet \cite{facenet}, etc. have come up, easily surpassing the performance obtained via hand-crafted methods with a great margin.

The rise in the performance in image recognition was observed along with the line of increasing depth of the CNN architectures such as GoogLeNet \cite{inception} and ResNet \cite{resnet}. Whereas, it is found that after certain depth, the performance tends to saturate towards mean accuracy, i.e., more depth has almost no effect over performance \cite{resnet}. At the same time, large scale application of face recognition would be prohibitive due to the need of high computational resources for deep architectures. Thus, in recent years, researchers are also working over the other aspects of the CNN model like loss functions, nonlinearities, optimizers, etc. One of the major works done in this field includes the development of suitable loss functions, specifically designed for face recognition. Early works towards loss functions include Center Loss \cite{centerloss} and Triplet Loss \cite{facenet} which focused on reducing the distance between the current sample and positive sample and increase the distance for the negative ones, thus closely relating to human recognition. Recent loss functions like Soft-Margin Softmax Loss \cite{softmargin}, Congenerous Cosine Loss \cite{coco}, Minimum Margin Loss \cite{minmarginal}, Range Loss \cite{rangeloss}, $L_2$-Softmax Loss \cite{l2loss}, Large-Margin Softmax Loss \cite{largemargin}, and A-Softmax Loss \cite{sphereface} have shown promising performance over lighter CNN models and some exceeding results over large CNN models.

Motivated by the recent rise in face recognition performance due to loss functions, this paper provides an extensive performance comparison of recently proposed loss functions for deep face recognition. Various experiments are conducted in this study to judge the performance of different loss functions from different aspects like effect of architecture such as deep and light weight and effect of training dataset. The results are analyzed using the training accuracy, test accuracy and rate of convergence. 
This paper is divided into following sections. Section \ref{lossfunctions} gives a comparative overview of the popular loss functions. Section \ref{netarch} describes the CNN architectures used. Section \ref{experimentsetup} discusses the training and testing setup. Section \ref{evals} presents the results. Section \ref{conclusion} concludes the paper.

\section{Loss Functions Used}
\label{lossfunctions}
As discussed in the earlier section, loss functions play an important role in CNN training. In this study, we discuss the widely used loss functions in face recognition. We have considered five loss functions, namely, Cross-Entropy Loss \cite{deepid2}, Angular-Softmax Loss \cite{sphereface}, Additive Margin Softmax Loss \cite{cosineface}, ArcFace Loss \cite{arcface}, and Marginal Loss \cite{marginal}. Some loss functions like Angular-Softmax Loss and Additive Margin Softmax Loss etc. are proposed specifically for the face recognition task. 

\textbf{Cross-Entropy Loss:}
The cross-entropy loss is one of the most widely used loss functions in deep learning for many applications \cite{goodfellow2016deep}, \cite{alexnet}. It is also known as softmax loss and has been proven quite effective in eliminating outliers in face recognition task as well \cite{parkhi2015deep}, \cite{deepid2}. The cross-entropy loss is given as,
\begin{equation}
\mathcal{L_{\textnormal{CE}}}=-\frac{1}{N}\sum_{i=1}^{N}\log\frac{e^{W^T_{y_i} x_i+b_{y_i}}}{\sum_{j=1}^{n}e^{W^T_j x_i+b_j}},
\label{eq:softmax}
\end{equation}
where $W$ is the weight matrix, $b$ is the bias term, $x_i$ is the $i^{th}$ training sample, $y_i$ is
the class label for $i^{th}$ training sample, $N$ is the number of samples, $W_j$ and ${W}_{y_i}$ are the $j^{th}$ and $y_i^{th}$ column of ${W}$, respectively.
The loss function has been used in the initial works done for face recognition tasks like the DeepID2 \cite{deepid2}, which has formed the foundation for current work in the domain.

\textbf{Angular-Softmax Loss:}
Liu et al. in 2017 published one of the many modifications to softmax function to introduce margin based learning. They proposed the Angular-Softmax (A-Softmax) loss that enables CNNs to learn angularly discriminative features \cite{sphereface}. It is defined as,
\begin{equation}
\mathcal{L_{\textnormal{AS}}}=-\frac{1}{N}\sum_{i=1}^{N}\log\big( \frac{e^{\|\bm{x}_i\|\psi(\theta_{y_i,i})}}{e^{\|\bm{x}_i\|\psi(\theta_{y_i,i})}+
\sum_{j\neq y_i}e^{\|\bm{x}_i\|\cos(\theta_{j,i})}} \big)
\label{angular}
\end{equation}
where $x_i$ is the $i^{th}$ training sample, $\thickmuskip=2mu \medmuskip=2mu \psi(\theta_{y_i,i})=(-1)^k\cos(m\theta_{y_i,i})-2k$ for $ \theta_{y_i,i}\in[\frac{k\pi}{m},\frac{(k+1)\pi}{m}]$, $\thickmuskip=2mu k\in[0,m-1]$ and $\thickmuskip=2mu m\geq1$ is an integer that controls the size of angular margin.
The performance of this function has been impressive, which has given a base for various margin based loss functions including CosineFace \cite{cosineface} and ArcFace \cite{arcface}.

\textbf{Additive-Margin Softmax Loss:}
Motivated from the improved performance of SphereFace using Angular-Softmax Loss, Wang et al. have worked on an additive margin for softmax loss and given a general function for large margin property \cite{cosineface}, described in following Equation,
\begin{equation}
\psi(\theta) = cos\theta - m.
\label{eq:psi_hardmargin}
\end{equation}
Using this margin, the authors have proposed the following loss function,
\begin{equation}
\begin{aligned}
\mathcal{L_{\textnormal{AM}}} & = -\frac{1}{N}\sum_{i=1}^N{\log\frac{e^{s \cdot  \left(cos\theta_{y_i} - m \right)}}{e^{s \cdot \left(cos\theta_{y_i} - m \right)} + \sum_{j=1,j\neq y_i}^{c}{e^{s \cdot  cos\theta_{j}}}}}\\.
\end{aligned}
\label{eq:am-softmax}
\end{equation}
where a hyper-parameter $s$ as suggested in \cite{cosineface} is used to scale up the cosine values.

\textbf{ArcFace Loss:}
Based on the above loss functions, Deng et al. have proposed a new margin $\cos(\theta+m)$ \cite{arcface}, which they state to be more stringent for classification. The angular margin \cite{arcface} represents the best geometrical interpretation as compared to SphereFace and CosineFace. The ArcFace Loss function using angular margin is formulated as,
\begin{equation}
\mathcal{L_{\textnormal{AF}}} =-\frac{1}{N}\sum_{i=1}^{N}\log\frac{e^{s \cdot (\cos(\theta_{y_i}+m))}}{e^{s \cdot\ (\cos(\theta_{y_i}+m))}+\sum_{j=1,j\neq  y_i}^{n}e^{s \cdot \cos\theta_{j}}},
\label{eq:aml}
\end{equation}
where $s$ is the radius of the hypersphere, $m$ is the additive angular margin penalty
between $x_i$ and $W_{y_i}$, and $cos(\theta + m)$ is the margin, which makes the class-separations more stringent. The ArcFace loss function has shown improved performance over the LFW dataset. Its performance is also very promising over the large-scale MegaFace dataset for face identification. 

\textbf{Marginal Loss:}
In 2017, Deng et al. proposed the Marginal Loss function \cite{marginal} which works simultaneously to maximize the inter-class distances as well as to minimize the intra-class variations, both being desired features of a loss function. In order to do so, the Margin Loss function focuses on the marginal samples. It is given as,
\begin{equation}
\mathcal{L}_{\textnormal{M}} =\frac{1}{N^{2}-N}\sum_{i,j,i\neq j}^{N}\bigg(\xi - {y}_{ij} \bigg( \theta - \Big\| \frac{x_i}{\|x_i\|} - \frac{x_j}{\|x_j\|} \Big\|^2_{2} \bigg) \bigg)
\label{eq:mrg}
\end{equation}
The term $y_{ij} \epsilon \{\pm1\}$ indicates whether the faces $x_i$ and $x_j$ are from the same class or not, $\theta$ is the distance threshold to distinguish whether the faces are from the same person or not, and $\xi$ is the error margin besides the classification hyperplane \cite{marginal}.
The final Marginal Loss function is defined as the joint supervision with regular Cross-Entropy (Softmax) Loss function and is given as,
\begin{equation}
\mathcal{L_{\textnormal{ML}}} = \mathcal{L}_{\textnormal{CE}} + \lambda\mathcal{L}_{\textnormal{M}}
\label{eq:mrg_final}
\end{equation}
where $\mathcal{L}_{\textnormal{CE}}$ is the cross-entropy (Softmax) Loss (Equation \ref{eq:softmax}). The hyper-parameter $\lambda$ is used for balancing the two losses. Usage of cross-entropy loss provides separable features and prevents the marginal loss from degrading to zeros \cite{marginal}.

\begin{figure*}[!t]
\centering
\includegraphics[width=0.8\linewidth]{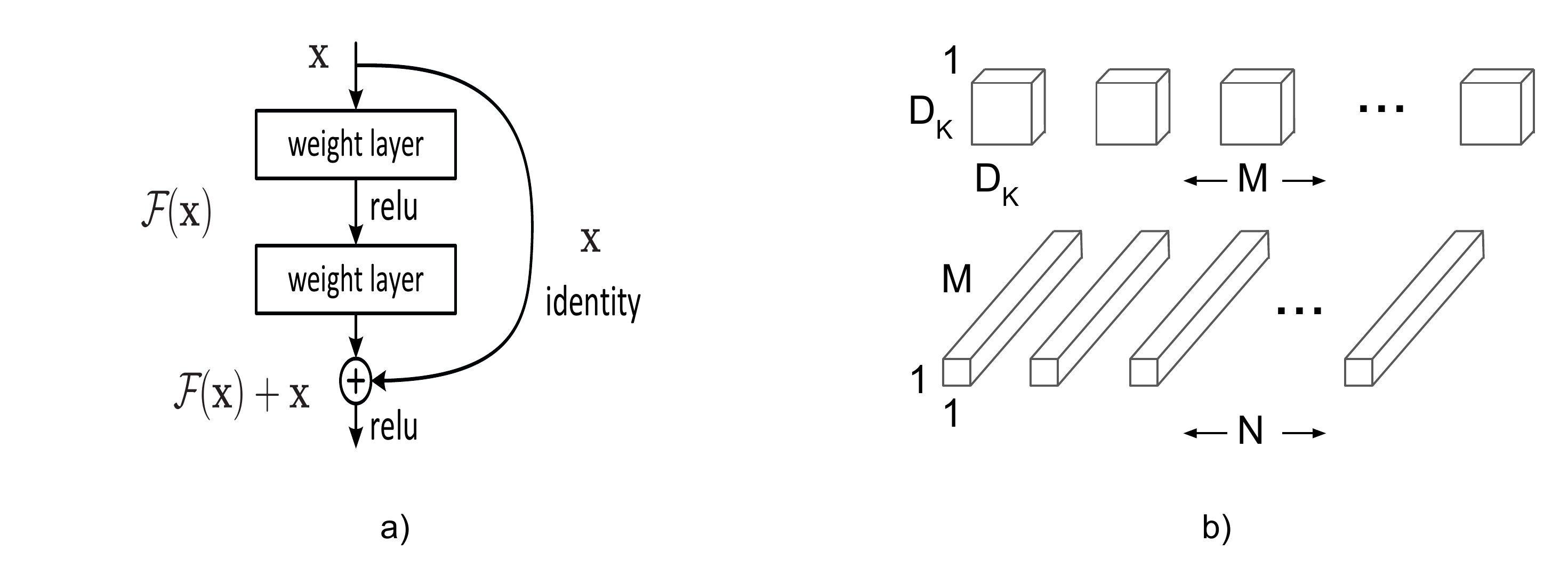}
\vspace{-6mm}
\caption{(a) Basic residual block used in ResNet \cite{resnet} which is a function where $X$ is the input and $F(x)$ is the function on $X$ and $X$ is added to the output of $F(X)$. (b) MobileNet uses two different convolution to reduce the computation. Here, $D_k$ is the filter size and $M$ is the input dimension. Next, $N$ filters of $1\times1$ dimension with $M$ depth are used to get output of the same dimension of $D_k$ output with depth on $N$. The Figures are taken from their respective papers.}
\label{fig:residual}
\end{figure*}

\section{Network Architectures}
\label{netarch}
\vspace{-0.2cm}
The CNNs have shown great performance for face recognition. We use the ResNet and MobileNet models to test over high performance and mobile platform scenarios.\\
\textbf{ResNet Model:}
\label{resnet}
\indent He et al. proposed a novel ResNet architecture which won the ImageNet challenge in 2015. The ResNet architecture is made with the building blocks of residual units. The Figure \ref{fig:residual}(a) demonstrate a residual unit. The ResNet unit learns a mapping between inputs and outputs using residual connections \cite{DBLP:Srivastava15}. This approach eliminates the problem of vanishing gradient as the identity mapping provides a clear pathway for the gradients to pass through the network. ResNet has proven to be a quite effective for a wide variety of vision tasks like image recognition, object detection and image segmentation. Hence, it makes the architecture one of the pioneer ensembles for face recognition tasks as visible from its many variants including ResNeXt \cite{resnext} and SphereFace \cite{sphereface}. In this paper, ResNet50 is used by keeping in mind the primary objective of evaluating efficiency of loss functions on standard architectures.\\
\textbf{MobileNet:}
\indent In 2017, Howard et al. presented a class of efficient architectures named MobileNets. These CNN models were designed with the primary aim of achieving efficient performance for mobile vision applications. This model uses the depth-wise separable convolutions as proposed by Chollet for the Xception architecture \cite{xception}. The building blocks for MobileNet are portrayed in Fig \ref{fig:residual}(b). The MobileNet architecture facilitates to build a light weight deep learning model. This model has shown the promising results over various vision based applications like object detection, face attributes and large scale Geo-localization with an efficient trade-off between latency and accuracy.

\section{Experimental Setup}
\label{experimentsetup}
\begin{center}
\begin{figure*}[!t]
\centering
\includegraphics[width=0.8\linewidth]{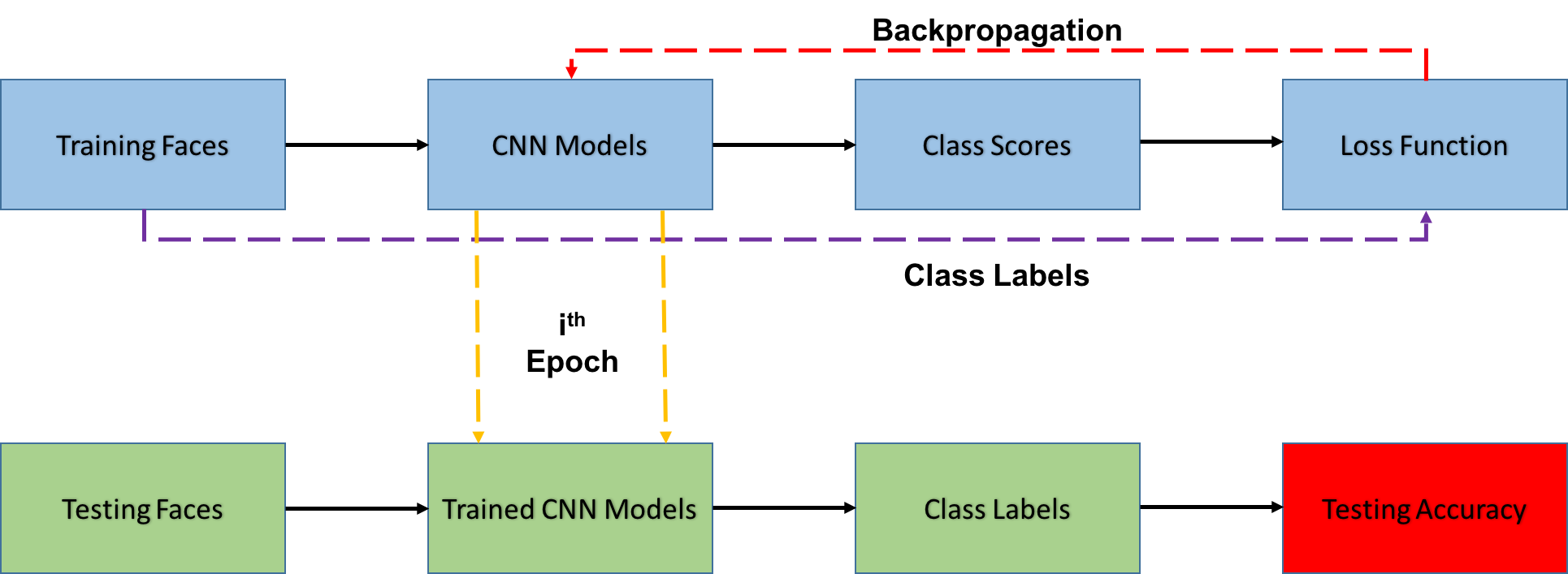}
\caption{Training and testing framework for performance evaluation of loss functions using CNN's. The $i^{th}$ epoch represents the transfer of the trained model after $i^{th}$ epoch's for testing.}
\label{fig:workflow}
\end{figure*}
\end{center}

\noindent\textbf{CNN based Face Recognition:}
The CNN based face recognition approach is illustrated in Fig. \ref{fig:workflow}. In each epoch, the learned weights obtained after training on all batches of training images are used to obtain the classification scores and accuracy over the training dataset. After training of each epoch, the trained weights at the moment are transferred to compute the accuracy over the test dataset.

\vspace{-1cm}

\noindent\textbf{Training Datasets:}
\label{tdata}
The CASIA-WebFace \cite{casia} is the most widely used publicly available face dataset. It contains 4,94,414 face images belonging to 10,575 different individuals. The original MS-Celeb-1M dataset \cite{msceleb} consists of 100k face identities with each identity having approximately 100 images resulting in about 10M images, which are scraped from public search engines. We have used a refined high-quality subset based on the clean list released by ArcFace \cite{arcface} authors. Finally, we obtained the MS-Celeb-1M dataset which contains 350k images with 8750 unique identities.\\
\textbf{Testing Dataset:}
Labeled Faces in the Wild (LFW) images \cite{lfw} are used as the testing dataset in this study. The LFW dataset contains 13,233 images of faces collected from the web. This dataset consists of the 5749 identities with 1680 people with two or more images. By following the standard LFW evaluation protocol \cite{lfw-report}, we have reported the verification accuracies on 6000 face pairs. \\
\textbf{Input Data and Network Settings:}
We have used MTCNN \cite{mtcnn} to detect facial landmarks to align the face images, similar to \cite{sphereface}, \cite{arcface}, \cite{marginal}. Each pixel in these images is normalized by subtracting 127.5 and then being divided by 128.
We have set the batch size as $64$ with the initial learning rate as $0.01$. The learning rate is divided by $10$ at the $8^{th}$, $12^{th}$ and $16^{th}$ epoch. The model is trained up to $20$ epochs. The number of epochs is less because the number of batches in an epoch is very high. The SGD optimizer with momentum is used for the optimization. The momentum and weight decay are set at $0.9$ and $5e^{-4}$, respectively.

\begin{figure*}[!t]
\centering
\includegraphics[width=\linewidth]{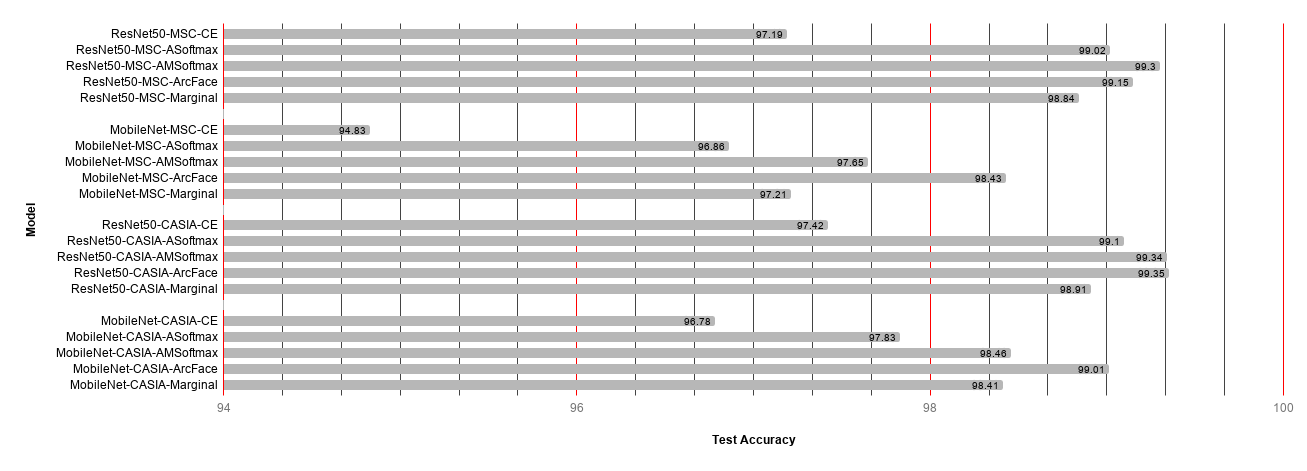}
\vspace{-1cm}
\caption{Highest test accuracies obtained over LFW dataset for different models under consideration in this study. These models have been trained on two datasets as described in the paper. The naming convention is as follows: Model Name-Training Dataset-Loss Function. Here, `CASIA' refers to CASIA-Webface and `MSC' refers to MS-Celeb-1M face datasets. For loss functions, `CE' refers to Cross Entropy loss, `ASoftmax' refers to the Angular Softmax loss and `AMSoftmax' refers to Additive Margin Softmax loss.}
\label{fig:test_acc}
\end{figure*}

\begin{figure*}[!t]
\centering
\includegraphics[width=\linewidth]{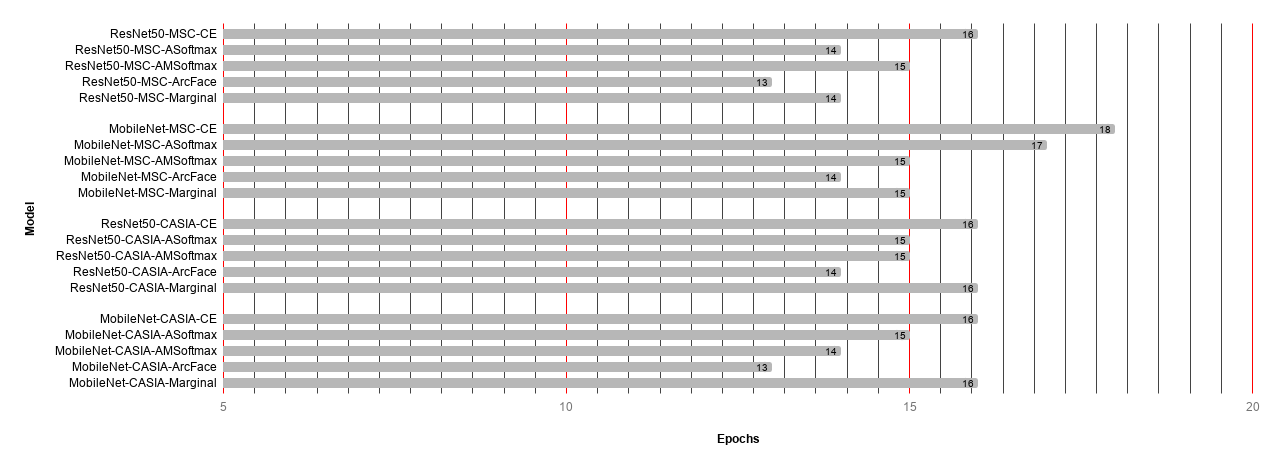}
\vspace{-1cm}
\caption{The minimum number of epochs taken to obtain the best model for a given loss function. The best model of a loss function gives the highest accuracy on LFW dataset. The naming convention is same as Figure \ref{fig:test_acc}.}
\label{fig:test_epoch}
\end{figure*}

\begin{center}
\begin{table*}[!t]
\caption{The training accuracies and testing accuracies obtained over LFW dataset performance comparison over two CNN architectures: ResNet50 and MobileNetv1 when trained on CASIA-Webface and MS-Celeb-1M datasets. The column `Training Accuracy' represents the accuracy obtained after training the model till 20th epoch. The term `Epochs' in the table signify the number of epochs at which we obtain the best model accuracy on LFW. The loss function AM Softmax refer to Angular-Margin Softmax. The last column, `Mean', denotes the mean of the test accuracies between the 10th and 20th epoch with the standard deviation in the same interval.}
\newcolumntype{Y}{>{\small\centering}p{0.15\linewidth}}
\newcolumntype{U}{>{\small\centering}p{0.19\linewidth}}
\newcolumntype{V}{>{\small\centering}p{0.20\linewidth}}
\newcolumntype{W}{>{\small\centering}p{0.08\linewidth}}
\newcolumntype{X}{>{\small\centering}p{0.09\linewidth}}
\newcolumntype{Z}{>{\small\centering}p{0.15\linewidth}}
\centering
\vspace{1mm}
\begin{tabular}{|Y|U|V|W|W|X|Z|}
\hline
\textbf{Base Model} & \textbf{Loss Function} & \textbf{Training Dataset} & \textbf{Train Acc} & \textbf{Test Acc} & \textbf{Epochs} & \textbf{Mean Accuracy} \tabularnewline
\hline
\hline
ResNet 50 & Cross Entropy & CASIA-Webface & 93.51 & 97.42 & 16 & $95.86\pm1.262$\tabularnewline
\hline
ResNet 50 & Cross Entropy & MS-Celeb-1M & 92.43 & 97.19 & 16& $95.84\pm1.254$\tabularnewline
\hline
\hline
ResNet 50 & Angular Softmax & CASIA-Webface & 94.01 & 99.10 & 15& $98.51\pm0.625$\tabularnewline
\hline
ResNet 50 & Angular Softmax & MS-Celeb-1M & 93.33 & 99.02 & 14 & $98.28\pm0.839$\tabularnewline
\hline
\hline
ResNet 50 & AM Softmax & CASIA-Webface & 94.37  & 99.34 & 15 & $98.65\pm1.044$\tabularnewline
\hline
ResNet 50 & AM Softmax & MS-Celeb-1M & 93.68  & 99.3 & 15 & $98.13\pm1.643$\tabularnewline
\hline
\hline
ResNet 50 & ArcFace & CASIA-Webface & 94.54 & 99.35 & 14& $99.01\pm0.305$ \tabularnewline
\hline
ResNet 50 & ArcFace & MS-Celeb-1M & 92.34 & 99.15 & 13& $98.06\pm1.532$ \tabularnewline
\hline
\hline
ResNet 50 & Marginal Loss & CASIA-Webface & 93.87 & 98.91 & 16 & $96.43\pm1.401$\tabularnewline
\hline
ResNet 50 & Marginal Loss & MS-Celeb-1M & 91.57 & 98.84 & 14 & $97.86\pm0.669$\tabularnewline
\hline
\hline
MobileNet v1 & Cross Entropy & CASIA-Webface & 93.42 & 96.78 & 16 & $95.59\pm1.030$\tabularnewline
\hline
MobileNet v1 & Cross Entropy & MS-Celeb-1M & 93.91 & 94.83 & 18 & $93.03\pm1.539$\tabularnewline
\hline
\hline
MobileNet v1 & Angular Softmax & CASIA-Webface & 92.47 & 97.83 & 15 & $96.34\pm1.120$\tabularnewline
\hline
MobileNet v1 & Angular Softmax & MS-Celeb-1M & 93.45 & 96.86 & 17 & $95.80\pm0.803$\tabularnewline
\hline
\hline
MobileNet v1 & AM Softmax & CASIA-Webface & 95.12 & 98.46 & 14 & $97.48\pm0.913$ \tabularnewline
\hline
MobileNet v1 & AM Softmax & MS-Celeb-1M & 94.10  & 97.65 & 15 & $96.47\pm1.165$\tabularnewline
\hline
\hline
MobileNet v1 & ArcFace & CASIA-Webface & 92.31 & 99.01 & 13 & $97.33\pm0.477$ \tabularnewline
\hline
MobileNet v1 & ArcFace & MS-Celeb-1M & 94.61 & 98.43 & 14 & $97.33\pm1.086$\tabularnewline
\hline
\hline
MobileNet v1 & Marginal Loss & CASIA-Webface & 93.15 & 98.41 & 16 & $97.10\pm1.428$\tabularnewline
\hline
MobileNet v1 & Marginal Loss & MS-Celeb-1M & 93.81 & 97.21 & 15 & $95.90\pm1.504$\tabularnewline
\hline
\end{tabular}
\label{table:observations}
\end{table*}
\end{center}

\section{Performance Evaluation and Observations}
\label{evals}
The loss functions as described in Section \ref{lossfunctions} are used with ResNet50 and MobileNetv1 CNN architectures to perform the training over MS-Celeb-1M and CASIA-Webface datasets and testing over LFW dataset. Here, we give a comparison of results based on test accuracies, rate of convergence and training and testing results. 

\subsection{Test Accuracy Comparison}
The different models have shown diverse performance when evaluated on the LFW dataset for face recognition. As evident from Figure \ref{fig:test_acc}, the two CNN architectures, ResNet50 and MobileNetv1 when trained on two face datasets, namely MS-Celeb-1M and CASIA-Webface show a varied performance of face recognition tasks. The best performing model obtained during the experiments is the ResNet50 model when trained on CASIA-Webface dataset using the ArcFace loss with an accuracy of 99.35\% on LFW dataset. The observed performance of ArcFace also resonates with its results when obtained with MobileNet architecture. It can be observed that the highest accuracies of 99.01\% and 98.43\% using MobileNetv1 are obtained using the ArcFace loss function when trained over CASIA-Webface and MS-Celeb-1M datasets, respectively. However, when ArcFace loss is used with ResNet50 and trained with the MS-Celeb-1M dataset, its accuracy of 99.15\% over LFW is slightly edged out by the Additive-Margin Softmax where we observed an accuracy of 99.30\%, the third best performing model obtained in our experiments.

In view of loss functions, the overall performance observed is in the following decreasing order: ArcFace, Additive Margin Softmax, Angular Softmax, Marginal Loss and Cross Entropy (Softmax). The Angular Softmax and Marginal loss almost have a similar performance with the first performing better with ResNet50 model while the latter showing better results with MobileNet model. The performance of Cross-Entropy Loss is not as good when compared to other losses. It can be justified as other four losses are proposed as the improvements over the Cross-Entropy Loss. The Additive Margin softmax Loss performed close to ArcFace Loss when observed with ResNet50 architecture, but lagged behind when MobileNet architecture is used. 

The performance difference observed for ArcFace Loss in ResNet and MobileNet architectures can be attributed to the base architecture itself. The ResNet 50 architecture used in our analysis is deep with 50 convolutional layers and residual modules. Whereas, the MobileNet architecture, has less number of convolutional layers and uses Depth Wise Separable Convolutions which tend to increase computation efficiency (for mobile devices) with certain tradeoffs. Moreover, the performance of other losses such as Angular Softmax, AM Softmax and Marginal loss is slightly lower than ArcFace due to the different ways of incorporating the margins.

Now coming to training datasets, we observed a distinct pattern when we evaluated models on LFW. The results that we obtained on both CNN architectures when trained on CASIA-Webface were comparatively better as compared to the same architectures trained on MS-Celeb-1M. One possibility for this observation stems out from the fact that MS-Celeb-1M contains more variations and even after extensive cleaning of the dataset as described in Section \ref{tdata}, there might be some existing noise as compared to CASIA-Webface.

\subsection{Convergence Rate Comparison}
In this paper, we define convergence rate in terms of the minimum number of epochs taken for a particular model to achieve it's highest test accuracy over LFW dataset. As discussed in the last section, a similar pattern of results is observed when we compare the convergence rate of loss functions for a same set of CNN architectures and training datasets. A comparison of convergence rate can be seen in Figure \ref{fig:test_epoch}. Again, the ArcFace loss has showed the fastest rate of convergence in all the models considered in this experiment. The ResNet architecture when trained on the MS-Celeb-1M dataset using ArcFace converged at $13^{th}$ epoch, the lowest epoch value seen in our tests. The same result is also observed with ArcFace when using MobileNet with CASIA-Webface training dataset.

Considering the two CNN architectures, ResNet50 and MobileNet, we observed a distinct pattern in terms of convergence rate when both the architectures are trained on the MS-Celeb-1M dataset. The ResNet model converged faster when compared to the MobileNet model with most of the loss functions, with the exception of Additive Margin Softmax Loss which converged on the 15th epoch for both the architectures. On the other hand, when the performance of architectures is observed over CASIA-Webface training dataset, a similar rate of convergence was observed for almost all the models based on ResNet and MobileNet using test dataset.

\subsection{Training and Testing Results Comparison}
The training and testing accuracies obtained during the experiments are summarized in Table \ref{table:observations}. The training accuracies reported in the table are obtained after training the model till the 20th epoch, that means after complete training of the model with the specified training dataset. Comparing the training accuracies, the highest accuracy of 95.12\% is obtained with the Additive Margin Softmax Loss when used with MobileNetv1 architecture and trained on CASIA-Webface dataset.

We have also computed the mean and standard deviation of testing accuracies obtained between the 10th and the 20th epoch to obtain the more generic performance of the loss functions discussed in Section \ref{lossfunctions}. These results also help us to observe the deviations of results between epochs as well as the convergence of the loss functions towards a saturation point. The highest mean accuracy of 99.01\% was observed for the ArcFace Loss when trained over CASIA-Webface dataset using the ResNet50 architecture with the standard deviation of 0.305. It is also observed that the above standard deviation is the lowest obtained over all the models considered in the experiments. Such a low standard deviation reaffirms the better performance of the ArcFace Loss over the epochs when compared to other loss functions discussed before in this study. The above observation resonates with other results like test accuracies and rate of convergence that we noticed in the previous sections, hence solidifying our computation of results obtained during the experiments.

\section{Conclusion}
\label{conclusion}
In this paper, we have presented a performance evaluation of recent loss functions with Convolutional Neural Networks for face recognition tasks. Recent loss functions like Angular-Softmax, Additive-Margin Softmax, ArcFace and Marginal Loss are compared and evaluated along with Cross-Entropy Loss. The ResNet50 and MobileNetv1 are used in our performance studies. Publicly available datasets like CASIA-Webface and MS-Celeb-1M are used for training the models. The performance is evaluated on the Labeled Faces in the Wild (LFW) dataset. The results are computed in terms of the training accuracy, test accuracy, and convergence rate. The ArcFace loss emerged as the best performing loss function with highest accuracy of 99.35\% over CASIA-Webface dataset. We evaluated the state-of-the-art losses for deep face recognition, which can help to the research community to choose among the different loss functions.

\section*{Acknowledgment}
This research is funded by Science and Engineering Research Board (SERB), Govt. of India under Early Career Research (ECR) scheme through SERB/ECR/2017/000082 project fund. We also gratefully acknowledge the support of NVIDIA Corporation with the donation of the GeForce Titan X Pascal GPU for our research.

{\small
\bibliographystyle{splncs04}
\bibliography{References}
}
\end{document}